\documentclass{article}
\usepackage{ijcai18}
\usepackage{times}
\usepackage{helvet}
\usepackage{courier}
\usepackage{algorithmic}
\usepackage{algorithm}
\usepackage{color}
\usepackage{graphicx}
\usepackage{multirow}
\usepackage{epsfig}
\def\mvp{\vspace*{-0.1in}}

\newcommand{\oimage}{\mathcal I}
\newcommand{\simage}{\widetilde{\mathcal{I}}}
\newcommand{\ovector}{\mathcal V}
\newcommand{\svector}{\widetilde{\mathcal{V}}}
\newcommand{\dsls}{\mathcal D}
\newcommand{\code}{\mathcal C}

\newcommand{\ignore}[1]{{}}

\begin{document}
\title{Automatically Generating Codes from Graphical Screenshots Based on Deep Autocoder}
\author{
Xiaoling Huang$^a$,
 and Feng Liao$^a$
\\
$^a$ Dept. of Computer Science, Sun Yat-sen University, Guangzhou, China
\\
liaof3@mail2.sysu.edu.cn,
}

\maketitle

\begin{abstract}
During software front-end development, the work to convert \emph{Graphical User Interface(GUI)} image to the corresponding front-end code is an inevitable tedious work. There have been some attempts to make this work to be automatic. However, the GUI code generated by these models is not accurate due to the lack of attention mechanism guidance. To solve this problem, we propose PixCoder based on an artificially supervised attention mechanism. The approach is to train a neural network to predict the style sheets in the input GUI image and then output a vector. PixCoder generate the GUI code targeting specific platform according to the output vector. The experimental results have shown the accuracy of the GUI code generated by PixCoder is over 95\%.
\end{abstract}

\section{Introduction}

Using machine learning techniques to build an automatic programming system is a relatively new field. Automatic programming can be applied to many fields. For example, Automatic programming can help developers to implement some simple functions. Balog et al. and Riedel et al. have done related work\cite{DBLP:journals/corr/BalogGBNT16,DBLP:journals/corr/RiedelBR16}. Balog et al. proposed DeepCoder, which take five input-output pairs as input, and output a program that satisfy these five input-output pairs. However, the program generated by DeepCoder is very simple, so Deepcoder is still far away from the actual application.

\begin{figure}[!ht]
  \centering
  \includegraphics[width=0.5\textwidth]{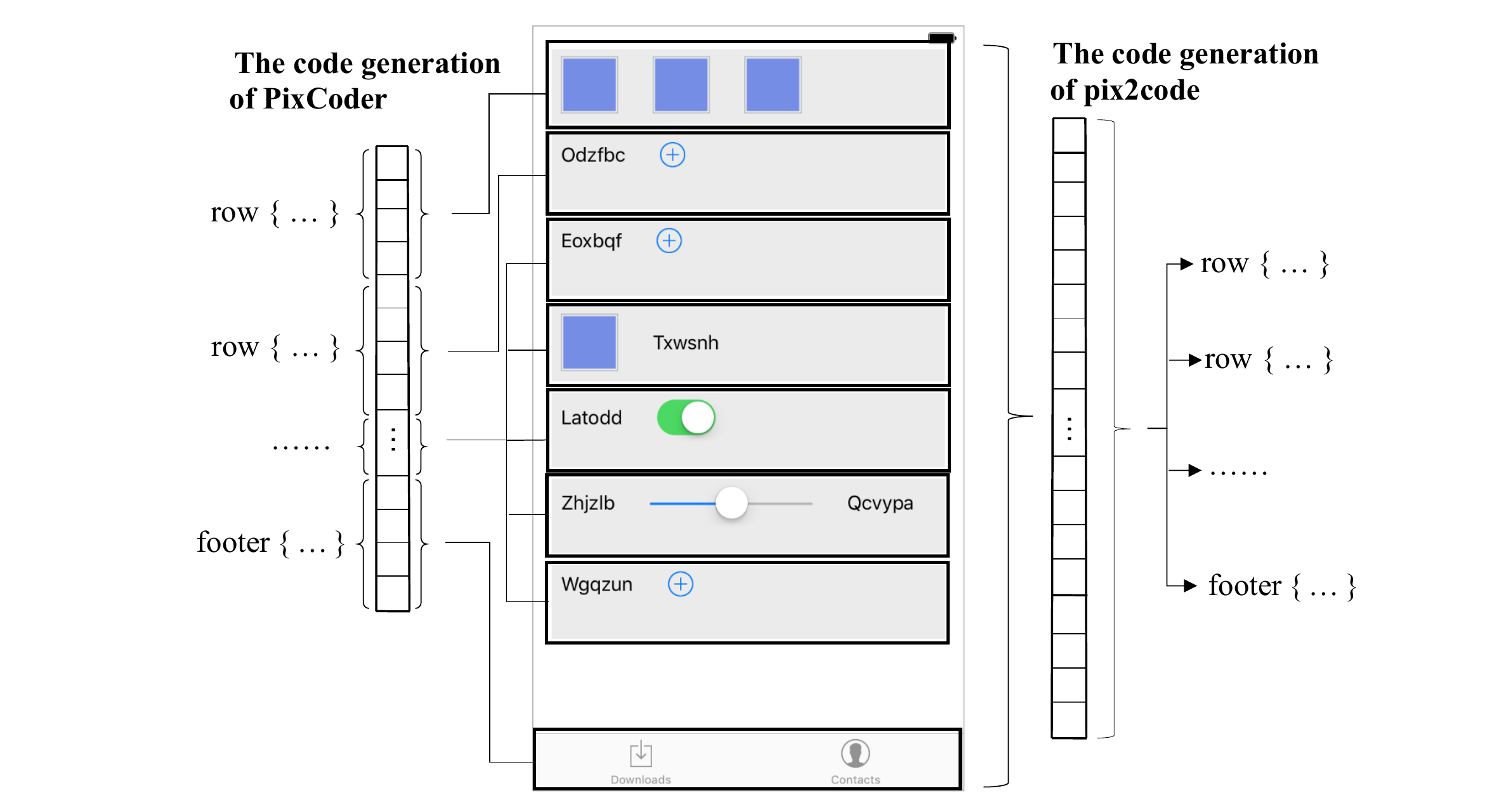}
  \caption{The code generation of PixCoder and pix2code.}
\mvp\mvp
\label{contrast}
\end{figure}

Moreover, automatic programming can prevent developers from time-consuming software front-end development. Beltramelli propose pix2code \cite{DBLP:journals/corr/Beltramelli17}, which is a system that takes a GUI image as input and outputs GUI code. Pix2code consists encoding part and decoding part. The encoding part contains a \emph{Convolutional Neural Network(CNN)} model and a RNN model. Pix2code accepts the GUI image and the corresponding DSLs as input. At this time, the CNN model is responsible for encoding the GUI image into a vector, and the RNN model is responsible for encoding a segment of the corresponding DSLs. And then input them to the decoding part. The decoding part is another RNN model. It generates the next character in the DSLs. By comparing the generated character with the character in the actual DSLs, a gradient can be obtained to optimize the parameters in pix2code. However, the result of pix2code is not satisfying since the code generation of pix2code lacks attention mechanism \cite{DBLP:journals/corr/LuongPM15}. As a result, the generation of every token is based on the whole image as shown in Figure \ref{contrast}. In fact, each piece of the frond-end code corresponds to a specific block in the image. This led to low accuracy of the front-end code generated by pix2code.

In this work, we focus on the problem in pix2code and propose PixCoder, which is a precise descriptive programming system using deep learning techniques. We apply the method named PixCoding, which applies an artificially supervised attention mechanism, to fix the problems in pix2code. In more detail, we first artificially encode the GUI image to a vector. As shown in Figure \ref{contrast}, Each bit in the vector is associated with some blocks in the image. In this way, we artificially establish the attention mechanism. As a result, the model generates code based on specific block of GUI image. And then We feed the images and vectors into a vision model based on CNN for supervised learning. During the training process, CNN learns the mapping between the style sheet in specific block of the GUI image and the specific bits in the vector. Finally, based on the predicted vector output by the vision model, we can directly generate the precise front-end code of GUI.


In summary, our concrete contributions are:
\begin{enumerate}
\item proposing an impressive automatic descriptive programming system. In some domains, the capacity of this system is comparable to humans.
\item defining a method in image recognition and classification. This method decomposes the original classification tasks into simple sub-classification tasks, which reduces the difficulty of the whole task and improves the classification accuracy.
\end{enumerate}

We organize the paper as follows. We first review related work. After that, we present the formal details of our framework, and then give a detailed description of our algorithm. Finally, we evaluate PixCoder in three different datasets and conclude our work with a discussion on future work.

\section{Related Work}
As we mentioned, there has been prior work on automatic programming. In automatic functional programming field, Bunel et al. propose an adaptive neural-compilation framework to address the problem of efficient program learning \cite{DBLP:conf/nips/BunelDMKT16}. In the work of Riedel et al., through a neural implementation of the dual stack machine that underlies Forth, programmers can write program sketches with slots that can be filled with behaviour trained from program input-output data \cite{DBLP:journals/corr/RiedelBR16}. Ling et al. explored the generation of source code from a mixed natural language and structured program specification \cite{DBLP:conf/acl/LingBGHKWS16}. In the work of Gaunt et al., the source code is generated through a distinguishable interpreter that learns the relationship between input-output examples \cite{DBLP:journals/corr/GauntBSKKTT16}. Balog et al. propose DeepCoder, which leverages statistical predictions to generate computer programs \cite{DBLP:journals/corr/BalogGBNT16}.

However, code generation with visual inputs is still an unexplored research area until Beltramelli propose pix2code \cite{DBLP:journals/corr/Beltramelli17}. The architecture of pix2code is similar to some models applied to other areas (c.f. \cite{DBLP:conf/cvpr/KarpathyL15,DBLP:conf/cvpr/VinyalsTBE15,DBLP:conf/icml/XuBKCCSZB15,DBLP:conf/cvpr/DonahueHGRVDS15}). Due to the code generation of pix2code lacks attention mechanism \cite{DBLP:journals/corr/LuongPM15}, the result of pix2code is far from expectation. For this problem, we apply an artificially supervised attention mechanism and adopt an approach similar to DeepCoder.

Most of these works rely on \emph{Domain Specific Languages(DSLs)}. DSLs are programming languages that are designed for a specialized domain. Compare with full-featured programming languages, DSLs are more restrictive. As a result, DSLs limit the complexity of programming language, which makes automatic programming easier and makes special-purpose search algorithm efficient \cite{DBLP:conf/oopsla/PolozovG15}. In this work, we are only interested in the GUI layout, the different graphical components, and their relationships. Thus we choice a simple DSLs designed by Beltramelli \cite{DBLP:journals/corr/Beltramelli17}.

\section{Problem Definition}
The problem we focus on is designing an automatic descriptive programming system, which input is a GUI image $\oimage$ and the output is the front-end code $\code$ of the GUI.
We denote $\alpha(\oimage)=\code$ as the automatic descriptive programming process. So the problem can can be considered to find mapping $\alpha$ between $\oimage$ and $\code$.

\begin{figure}[!ht]
  \centering
  \includegraphics[width=0.425\textwidth]{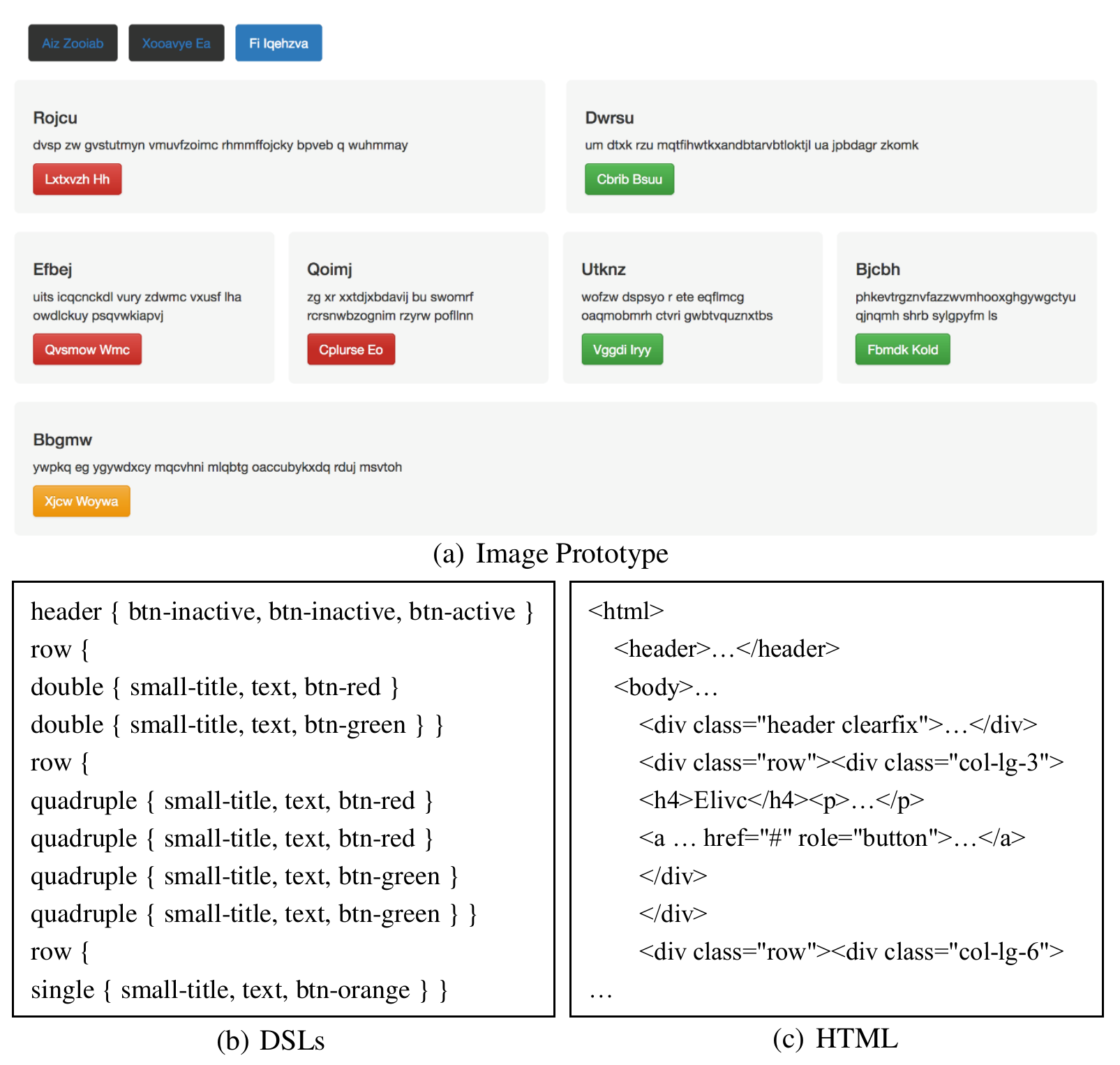}
  \caption{An example of GUI image, DSLs and front-end code.}
\mvp\mvp
\label{imgdslcode}
\end{figure}

According to different target platforms, $\oimage$ can be a web-based UI image, an iOS UI or an Android UI image. Figure \ref{imgdslcode}(a) is an input example of web-based UI image. Different UI images vary in size, so we resize the input image to 256$\times$256 pixels and the pixel values are normalized. After this process, we get standardized image $\simage$.

The output of the problem is the front-end code $\code$ of the GUI. However, as shown in Figure \ref{imgdslcode}(c), the front-end code is complicated. And it is difficult for the model to directly generate the front-end code. So our approach is designing DSLs $\dsls$ for target front-end code $\code$ and let model output $\dsls$ instead of $\code$. Figure \ref{imgdslcode}(b) is the DSLs of web-based UI image. The syntax of DSLs is relatively simple here, since we only need to use $\dsls$ to describe the outline of the GUI image.

Since the output of our model is $\dsls$ and the problem need $\code$. So we need to compile $\dsls$ to get $\code$. This process is denoted as $\beta(\dsls)=\code$ and we denote $\delta(\simage)=\dsls$ as the automatic descriptive programming process of model. The compile process is static so the key work is to find $\delta$, the mapping between $\simage$ and $\dsls$.

\section{PixCoder}
To implement such an automatic descriptive programming system, we have tried to add attentional mechanism to pix2code just like the work of Luong et al. \cite{DBLP:journals/corr/LuongPM15}. However, this did not improve the performance of the model. Therefore, we use a method named PixCoding to add a supervised attentional mechanism artificially. And in combination with the ideas of DeepCoder \cite{DBLP:journals/corr/BalogGBNT16}, we propose PixCoder. The automatic descriptive programming process of PixCoder can be divided into the following phases: The first phase is a vision model based on CNN, which is used to understanding the given GUI image $\oimage$ and inferring the objects present. The vision model outputs a predicted vector $\ovector$ and each bit in $\ovector$ corresponds to a kind of style sheet in different block in $\oimage$. This can be seen as a vision model identifies each block in $\oimage$ and then record the result into $\ovector$. In the second phase, we exploit a parser in solving code generation problem. The parser leverages $\ovector$ generated by vision model to generate DSLs $\dsls$ describing $\oimage$. The last phase is a compiler. In this phase, with traditional compiler design techniques, $\dsls$ is compiled to the front-end code $\code$ targeting specific platform.

\begin{figure}[!ht]
  \centering
  \includegraphics[width=0.4\textwidth]{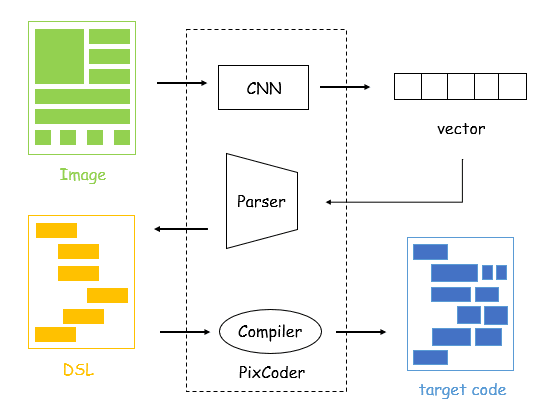}
  \caption{PixCoder model architecture. PixCoder consists of vision model, parser and compiler.}
\mvp\mvp
\label{architecture}
\end{figure}

\subsection{PixCoding}
PixCoding is a universal image recognition method: we analyze the possible changes in different areas of the image and then devise a vector. Each bit in the vector corresponds to a different change in different areas of the image. And then we can make some rules based on our experience, according to which we are able to know that the changes in various combinations of different areas corresponds to which label. And then the task of image recognition can be completed. This image recognition method decomposes the image classification task into sub-classification tasks in different areas of the image, and then combines the results of the sub-classification tasks to obtain the final classification result. At this point, the accuracy of the classification depends on the accuracy of the sub-classification. Because sub-tasks are relatively simple, this method can improve the accuracy of the whole task.

In this work, we apply this method to automatic descriptive programming field. We design the corresponding vectors for a particular type of GUI image. Each bit in the vector corresponds to the changes of style sheet in different blocks in the GUI image. By combining each of the vectors we know all the style sheets in the GUI image. As a result, we can directly generate DSLs corresponding to the input GUI image. This process is no different from the process of combining the results of a sub-classification task to get the final classification result. In addition, PixCoding mentioned above can be applied to image text recognition, such as recognition of long strings. We only need to identify the different areas of the characters belong to which number, and then combined them to complete the string recognition task. This method can also be applied to motion recognition. We divide the hand, foot, torso and other parts of human into different areas and then encode the vectors so that the vectors imply all possible situations in these areas.

However, PixCoding has some limitations: it can only be used when the data in the dataset is very normative and has a clearly defined area.

\begin{algorithm}[!ht]
\caption{PixCoder}\label{algorithm:main}
\textbf{Input:} a GUI image $\oimage$. \\
\textbf{Output:} the GUI's front-end code $\code$.

\begin{algorithmic}[1]
\STATE resize $\oimage$ to 256$\times$256 pixels and normalized the pixel values. And get standardized image $\simage$: \emph{Image\_Standardization}($\oimage$)=$\simage$;
\STATE input $\simage$ to vision model and get predicted vector $\ovector$: $\zeta(\simage)=\ovector$;
\STATE standardize $\ovector$ and get standardized vector $\svector$: \emph{Vector\_Standardization}($\ovector$)=$\svector$;
\STATE generate DSLs $\dsls$ according to $\svector$: $\theta(\svector)=\dsls$;
\STATE compile $\dsls$ to get $\code$: $\beta(\dsls)=\code$;
\RETURN $\code$;
\end{algorithmic}
\end{algorithm}

\subsection{Image To Vector}

\subsubsection{Vector Design}
\begin{figure}[!ht]
  \centering
  \includegraphics[width=0.4\textwidth]{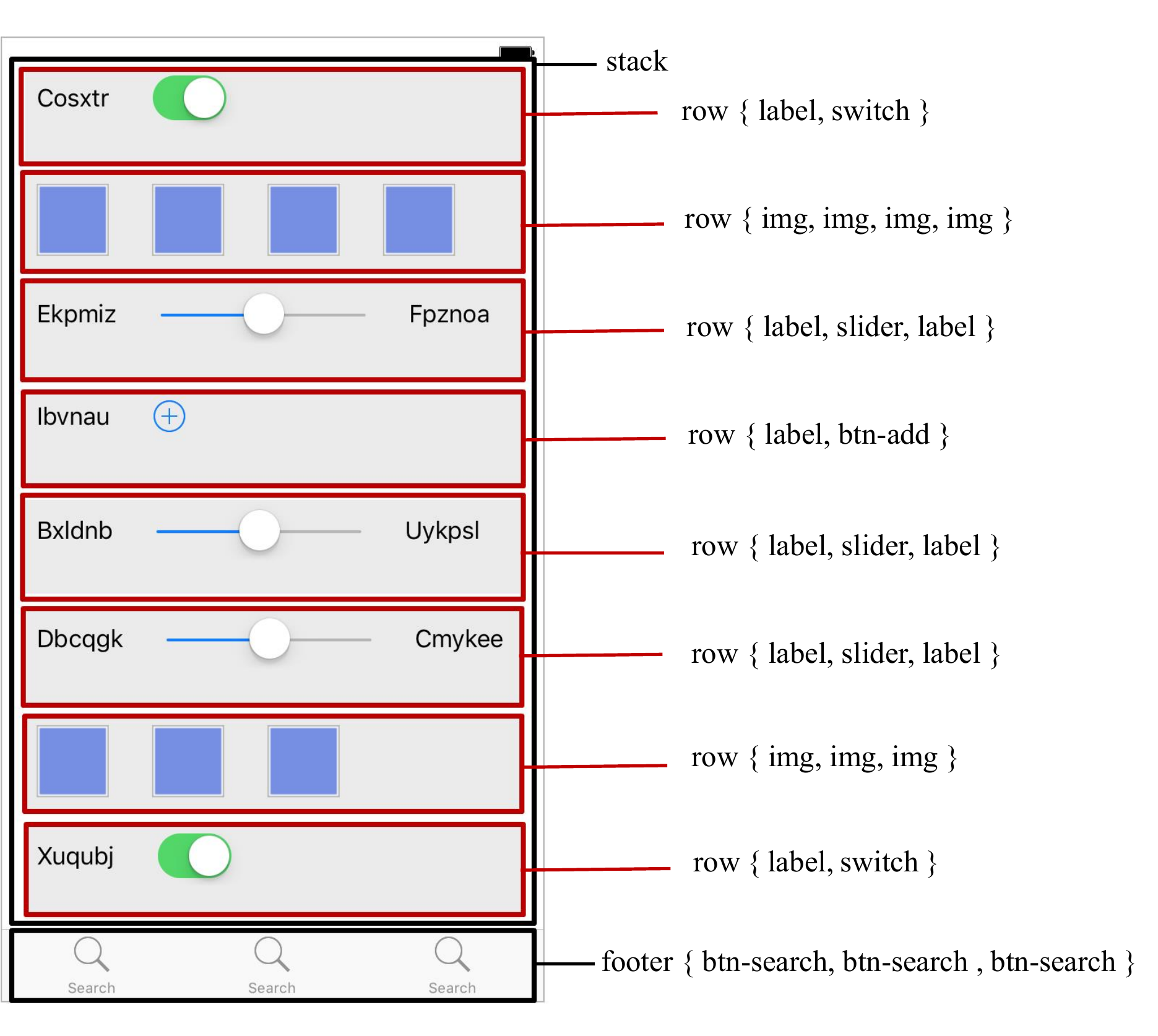}
  \caption{GUI image can be divided into several blocks.}
\mvp\mvp
\label{blockinimg}
\end{figure}

First, we need to analyze and design a specific type of GUI image. An GUI image example from iOS UI dataset is shown in Figure \ref{blockinimg}. After analyzing the iOS UI dataset, we learned that the GUI image here can be divided into two blocks: a stack block and a footer block (the area in the Figure \ref{blockinimg} is surrounded by the black frame). There can be a maximum of eight row blocks in the stack block, and at least one row block (the area in the Figure \ref{blockinimg} that is surrounded by the red frame). There are at most four controls in each row block, and there are five types of controls, and up to four controls in the footer block and four types of footer controls. If all possibilities are under consideration, we shall design a vector of 176 (4$\times$5$\times$8+4$\times$4) bits using one-hot encoding. In this case, the vector contains all the blocks and the changes of the style sheet of them and in the GUI image.

\begin{figure}[!ht]
  \centering
  \includegraphics[width=0.405\textwidth]{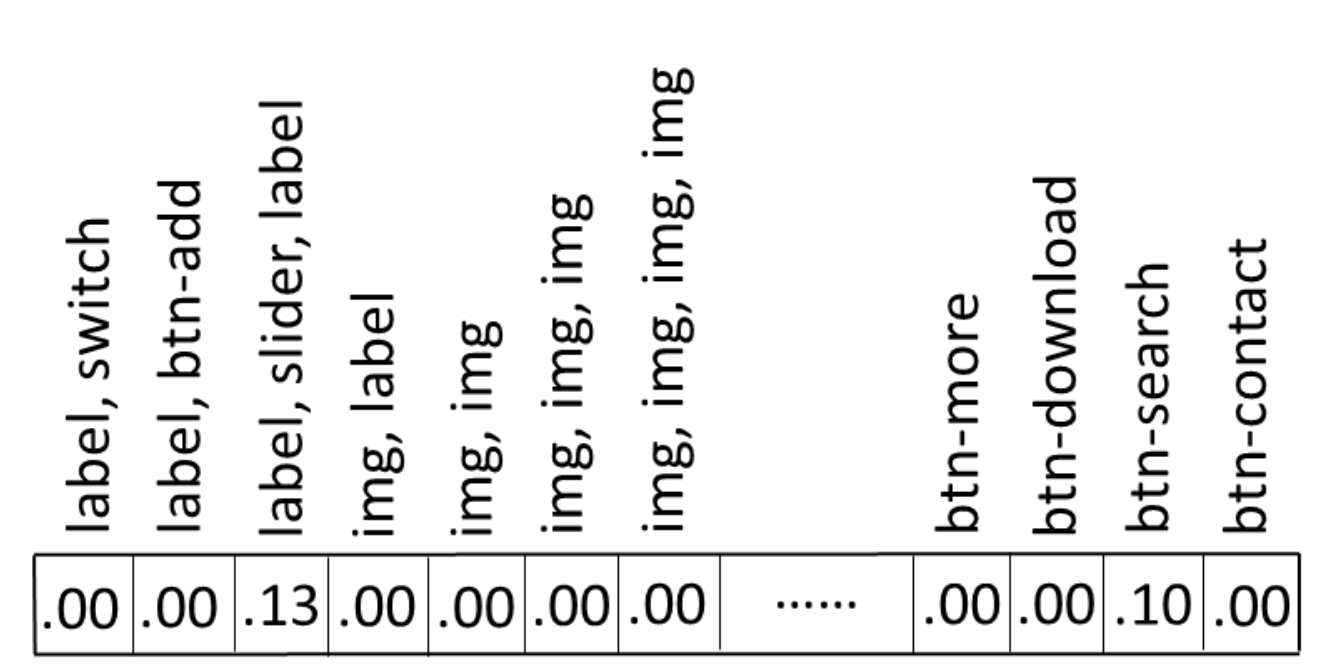}
  \caption{The vector we design for iOS UI image.}
\mvp\mvp
\label{vector}
\end{figure}

But if we do further analysis, we can prune the bits of the vector. After further analysis, we know that the number of row blocks is from 1 to 8; the number of controls in the row block is from 2 to 4. The number of controls in the footer block ranges from 2 to 4. The appearance of controls in the row block is somewhat regular: such that the slider does not appear alone, it always appears as a group like label, slider, label. According to the analysis results, we can prune the median of the vector, and finally get a 72-bit vector shown in Figure \ref{vector}.

Using the vector before and after pruning separately, we found that the use of pruned vectors can reduce training time and improve the accuracy of vision recognition. In addition, the pruned vector can also avoid some unusual circumstances. Such as the error of beyond the length: Due to the length of each control is different, so a row block cannot contain four longest slider controls. In the pruned vector, the situation which four sliders appear in a row block cannot happen. As a result, we can avoid this error of beyond the specified length, which in turn enhances our automatically descriptively programming system.

\subsubsection{Vision Model}
After designing the vector, we need a vision model to achieve the image-to-vector mapping $\zeta$. we employ a CNN to model and learn $\zeta$ due to it can learn rich latent representations from the image \cite{DBLP:conf/nips/KrizhevskySH12,DBLP:journals/corr/SermanetEZMFL13}. CNN is one of the most widely used methods in vision problems. Compared with the traditional image processing algorithms, CNN avoids the complex pre-processing of images, which improves the efficiency of the experiment greatly.

Firstly, $\oimage$ are initially re-sized to 256$\times$256 pixels and normalized the pixel values. After this process, we get standardized image $\simage$ and fed it in CNN. To recognize different objects in input image, we exclusively used 3$\times$3 receptive fields which are convolved with stride 1 as VGGNet \cite{DBLP:journals/corr/SimonyanZ14a}. These operations are applied twice before to down-sample with max-pooling. The width of the first convolutional layer is 32, followed by a convolutional layer of width 64, and the width of the third convolutional layeris 128. Our CNN ends with two fully connected layers. The size of the first fully connected layer is 1024 and the size of the final connection layer is based on the complexity of the style sheets in given GUI images. Except for the final fully connected layer applying the softmax activation function, each layer of our model applying the rectified linear unit activation function.

The Training process of vision model is supervised. Since only $\oimage$ and $\dsls$ are provided in the dataset, after designing the vector, we need to generate $\ovector$ corresponding to $\oimage$ and use it as the label for the supervised training. After completing this pre-process, we can start to train the vision model.

\begin{table}[!ht]
\mvp
\caption{Patterns of GUI image statistics.}\label{Number of all possible patterns}
\centering
  \begin{tabular}{|c|c|c|}
  \hline
  \multirow{2}{*}{\textbf{Dataset}} & \multicolumn{2}{c|}{\textbf{Number of all possible patterns}}\\
  \cline{2-3}
   & Before pruning & After pruning\\
  \hline
  web-based UI Image & 1.08$\times$$10^{8}$ & 3528\\
  \hline
  iOS UI Image & 4.66$\times$$10^{25}$ & 9072\\
  \hline
  Android UI Image & 1.16$\times$$10^{28}$ & 42768\\
  \hline
  \end{tabular}
\end{table}

\begin{figure}[!ht]
  \centering
  \includegraphics[width=0.405\textwidth]{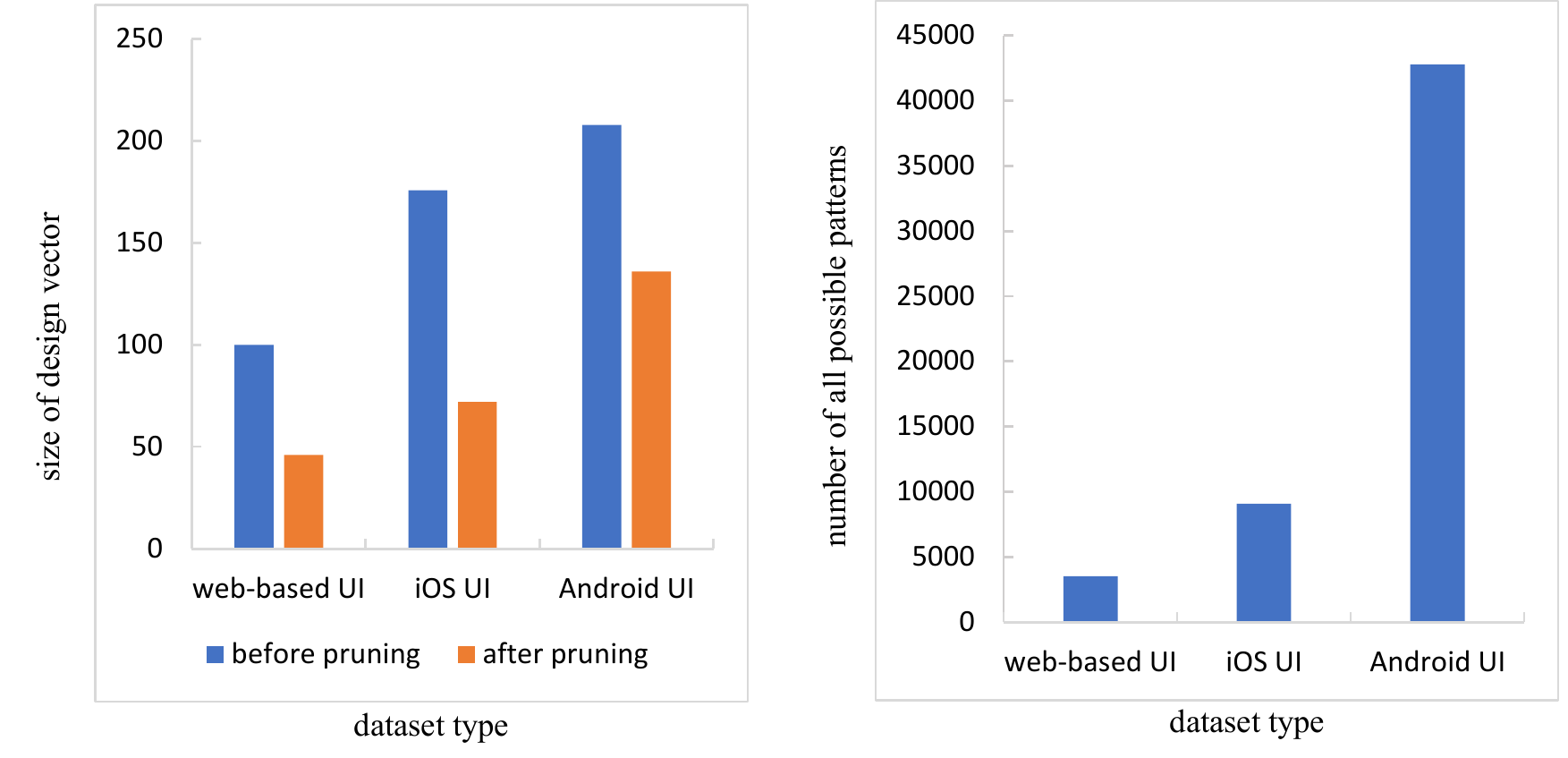}
  \caption{The vector's size before and after pruning statistics and the patterns of different GUI image statistics.}
\mvp\mvp
\label{varietyofvectorsize}
\end{figure}

\subsection{Vector To DSLs}
The central idea of our work is to use a vision model to guide code generation. The parser is responsible for generating DSLs based on vectors. The parser has two main functions: to standardize $\ovector$ output by the vision model, and to generate $\dsls$ based on the $\svector$($\theta(\svector)=\dsls$).

As mentioned above, the vector we design contains all the stylistic changes in the GUI image. Therefore, we can know from the vector that the presence or absence of individual style sheets in individual block of the given input GUI image. In accordance with the previously designed DSLs syntax rules, we can generate the corresponding DSLs directly from the output vector.

However, $\ovector$ output by the vision model does not meet our expectations. First, the value of each bit in the output vector is not a standard value like 0 or 1. In addition, it is possible to have ambiguities in the output vectors, such as specifying conflicting patterns for a block, and so on. Therefore, before generating the corresponding DSLs based on vectors, we need to standardize $\ovector$ and get $\svector$. We can only generate $\dsls$ based on $\svector$.

\subsubsection{Vector Standardization}
First, we need to use thresholds to process the value of each bit in $\ovector$ into a standard value like 0 or 1. Here we cannot use the method of setting the maximum value in each region of $\ovector$ directly to 1, and the rest of the value all to 0. Because this method does not consider the situation that $\ovector$ can have some regions of all 0. Using the iOS UI dataset as an example, with our method, for any input GUI image, the final output is a DSLs containing eight row blocks. Therefore, we need to use thresholds to deal with this problem.

In this work, we automatically get the threshold during the training of vision model. After each epoch after training halfway, we take a part of the samples from the training set and input them into the vision model to get the output vector. By comparing the output vector with the target vector, we can know which bit in the output vector correspond to 1 and which bit correspond to 0. The smallest one of the output values corresponding to 1 is recorded as a candidate for the threshold value. At the same time, we will also record the largest of the output values corresponding to 0. After training, We will apply a complex algorithm to get the threshold. For example, if the maximum value corresponding to 0 is smaller than the minimum one corresponding to 1, then we take the average of this interval as the threshold. Otherwise, We use other methods to get a reasonable threshold. According to Figure \ref{accthreshold}, The choice of threshold is important to the result of model. And the threshold generated by our algorithm is close to the best threshold.

For all possible ambiguities in $\ovector$, we have corresponding solution. For example, for the situation of specifying conflict patterns for a block, our approach is to take the larger one of the two models corresponding to the prediction and set it to 1, and the other set to 0. And for the situation that an empty row block in the middle of two row blocks, our approach is to swap the empty row block with next row block contains objects.

\subsubsection{DSLs Generation}
After completing the process of standardizing $\ovector$, we can easily generate the corresponding DSLs according to $\svector$ based on the previously designed DSLs grammar rules($\theta(\svector)=\dsls$). The normalized vector is error-free, which guarantees there will be no syntax error exists in the generated DSLs.

\subsection{DSLs to Front-end code}
When we get $\dsls$ from parser, our work have not finished. We need a compiler to compile the $\dsls$ into $\code$ for various platforms(i.e. Android and iOS native mobile interfaces, and multi-platform web-based HTML/CSS interfaces). As mentioned, This process denotes as $\beta(\dsls)=\code$. The compile process is static. Thanks to the parser guarantees $\dsls$ without syntax error, $\code$ generated by compiler is legal.

\section{Experiments}
\subsection{Dataset}
The dataset provided by Tony Beltramelli \cite{DBLP:journals/corr/Beltramelli17} has three types, including iOS GUI images and corresponding DSLs, Android GUI images and corresponding DSLs and web-based GUI images and corresponding DSLs. Each type of dataset is divided into two parts, including 1500 training set and 250 testing set. Each example contains an original image and the corresponding DSLs.

According to Figure \ref{varietyofvectorsize}, the complexity of these three datasets is increasing. The number of the style sheets of the GUI image are least in the Web-based UI dataset, the iOS UI dataset is the second, and the Android UI datasets is the most.

\begin{table}[!ht]
\mvp
\caption{Dataset statistics.}\label{dataset}
\centering
  \begin{tabular}{|c|c|c|c|c|}
  \hline
  \multirow{2}{*}{\textbf{Dataset}} & \multicolumn{2}{c|}{\textbf{Training set}} & \multicolumn{2}{c|}{\textbf{Test set}}\\
  \cline{2-5}
   & Image & DSLs & Image & DSLs\\
  \hline
  web-based UI & 1500 & 1500 & 250 & 250\\
  \hline
  iOS UI & 1500 & 1500 & 250 & 250 \\
  \hline
  Android UI & 1500 & 1500 & 250 & 250 \\
  \hline
  \end{tabular}
\end{table}


\subsection{Criterion}
There are many ways to evaluate the quality of codes generated by the model. For example, screenshot the image showed by the generated GUI code, and then compare the similarities between this and the input image. Or, compare the similarities between the generated code and the target code. After experimental comparison, we evaluated the PixCoder with the second way: compare the similarities between the generated DSLs and the target DSLs.

In our experiment, every DSLs that in line with grammatical rules is in tree structure. We first convert the generated DSLs and the target DSLs to a tree, then using the similarities between tree structures to represent the similarities between DSLs. The classic algorithm used to match the similarity of traditional tree models is based on the similarity of edit distance \cite{DBLP:journals/jacm/Tai79}, which allows cross-layer matching of tree nodes. Cross-layer matching and replacement may be useful when comparing the general tree similarity, but not suited to the tree model used in our experiment. The tree model of our experiment is similar to the HTML tree model. HTML tree model document label nodes will be read by the browser rendering to the screen, different root node corresponds to a different set of child nodes. Therefore, even if we replace the root node, the children will not be matched. So cross-layer matching and replacement cannot obfuscate the relationship between nodes in the HTML tree, which is rendered in the browser effect.

Therefore, we need another way to measure the similarity of the tree model in our experiment. Here we use the Simple Tree Matching(STM) algorithm \cite{DBLP:journals/spe/Yang91,DBLP:journals/jcrd/He07}. STM bases on the principle of maximum matching, dynamic programming is used to calculate the maximum number of matching nodes of two trees, and then the similarity between two trees is obtained. This process does not allow cross-layer matching and node replacement, and requires that child nodes be completely ordered. If the root nodes of two subtrees do not match, the other nodes of the two subtrees are not considered, so as to achieve the effect of pruning. The time complexity of this algorithm is~O($n^{2}$).

$$Similarity(T_1, T_2) = \frac{SimpleTreeMatching(T_1, T_2)}{(|T_1|+|T_2|)/2}$$

Above is the formula that calculates the similarity between the generated DSLs and the target DSLs. ~$T_{1}$~and~$T_{2}$~represent the tree obtained by converting the generated DSLs and the target DSLs.

SimpleTreeMatching($T_{1}$, $T_{2}$)~represents the maximum number of matched nodes for two trees. $|T_{1}|$~and~$|T_{2}|$~represent the number of nodes in two trees. If and only if the maximum number of matching nodes of two trees is larger, the Similarity($T_{1}$, $T_{2}$)~is larger, too. That is, SimpleTreeMatching($T_{1}$, $T_{2}$)~is closer to 1. At this point the trees represent the generated DSLs and the target DSLs are more similar.

\begin{table}[!ht]
\mvp
\caption{Experimental results in different dataset.}\label{result}
\centering
  \begin{tabular}{|c|c|c|c|}
  \hline
  \multirow{2}{*}{\textbf{Model}} & \multicolumn{3}{c|}{\textbf{Dataset}}\\
  \cline{2-4}
   & web-based UI & iOS UI & Android UI\\
  \hline
  Baseline & 62.882\% & 70.303\% & 65.825\%\\
  \hline
  pix2code(beam 3) & 76.905\% & 68.640\% & 54.644\%\\
  \hline
  pix2code(greedy) & 88.591\% & 87.621\% & 85.073\%\\
  \hline
  PixCoder & 98.699\% & 95.562\% & 98.177\%\\
  \hline
  \end{tabular}
\end{table}

\begin{figure}[!ht]
  \centering
  \includegraphics[width=0.405\textwidth]{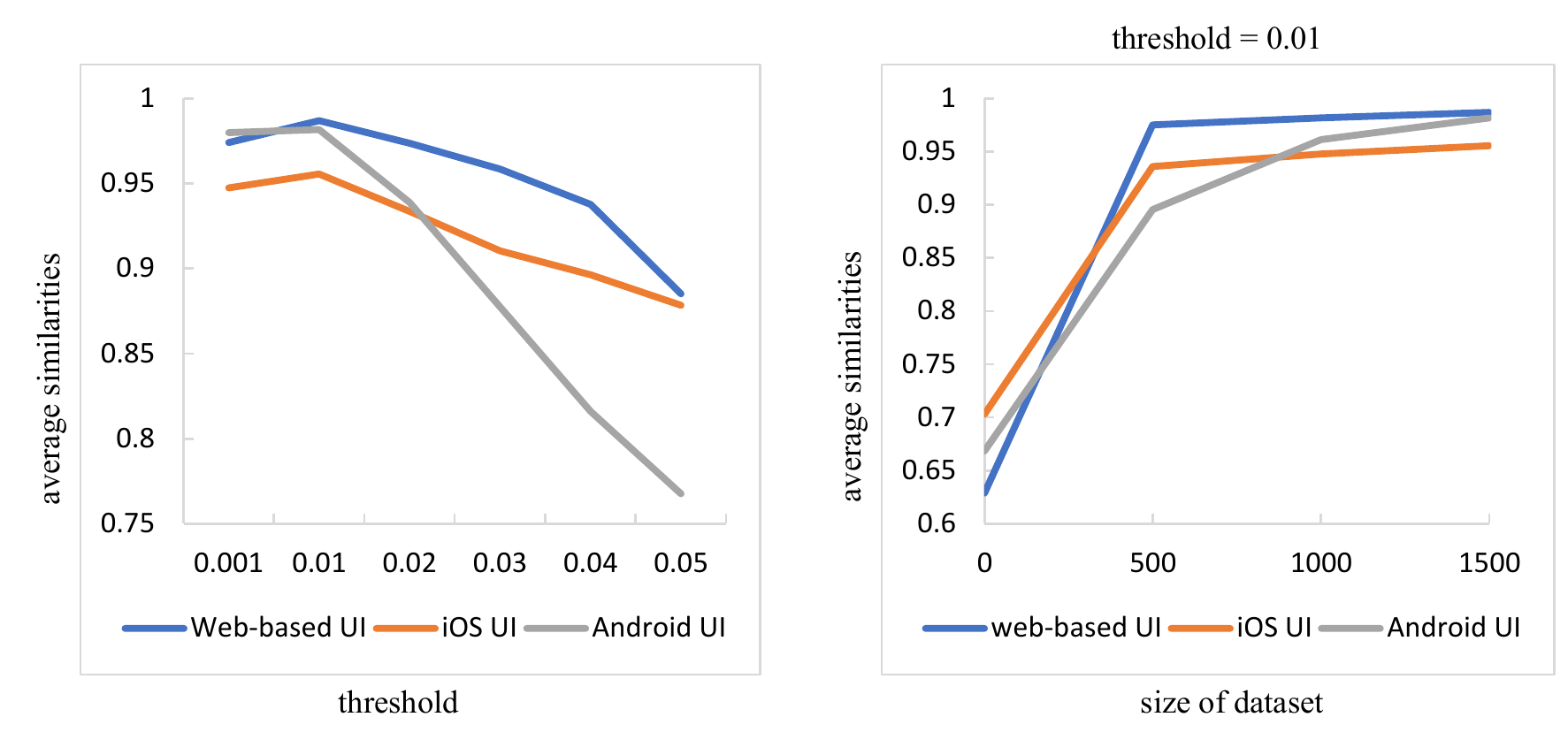}
  \caption{The line chart of PixCoder's average similarities varies with threshold and dataset's size.}
\mvp\mvp
\label{accthreshold}
\end{figure}


\subsection{Experimental Results}
We evaluate PixCoder in following aspects. We first compare the similarities between the target DSLs and the DSLs generated by PixCoder and Baseline. We then compare the similarities between the target DSLs and the DSLs generated by PixCoder and pix2code, specially, pix2code has two types, one in use of beam 3 method, and the other in use of greedy method. We use a randomly generated grammar-compliant DSLs with a similarity to the target DSLs of the test set as Baseline.

As shown in Figure \ref{accthreshold}, we can get the conclusion that the choice of threshold is important to the result of PixCoder. Threshold can not be too large or too small. In this experiment, we set threshold to 0.01, which is close to the best threshold according to the experimental results. As shown in Figure \ref{accthreshold}, the size of the dataset affects the result. As the dataset increases, the model can learn more knowledge about the style sheets of GUI image. As a result, the model is more capable. In this work, we feed all 1500 examples from dataset to the model while training.

The experimental results in the Table \ref{result} are the average similarities of 250 samples in the test set obtained using baseline, pix2code and PixCoder respectively. As described above, the complexity of these three datasets is increasing. The average similarities of Baseline on these datasets do not reflect the complexity of the style sheet changes in the GUI image. We believe this is caused by the unevenly distributed style sheets of the GUI image in the test datasets. The main contrast model of PixCoder in this experiment is pix2code. Constrained by the experimental results, the DSLs generated by pix2code in use of beam 3 method has a lot of syntax errors. In the calculation of average similarities, our method is to write 0 for the similarity of DSLs that have syntax errors. This results in the similarities of pix2code in use of beam 3 method being particularly low. Pix2code in the use of the greedy method almost has no syntax errors, so its average similarities are relatively high. There is no syntax error in DSLs generated by PixCoder, and its accuracy is far better than pix2code. The difference in average similarities between Pix2code and PixCoder does not reflect the difference in the quality of the code they generate. In fact, the quality of the code generated by Pix2code is far from PixCoder. As shown in Figure \ref{result}, There is only one error(the area surrounded by the red frame) in the output of PixCoder and there are eight error in pix2code. But their difference in similarity is only 12.821\%. In a word, to the best of our knowledge, PixCoder is the best model in automatic descriptive programming field.

\begin{figure}[!ht]
  \centering
  \includegraphics[width=0.45\textwidth]{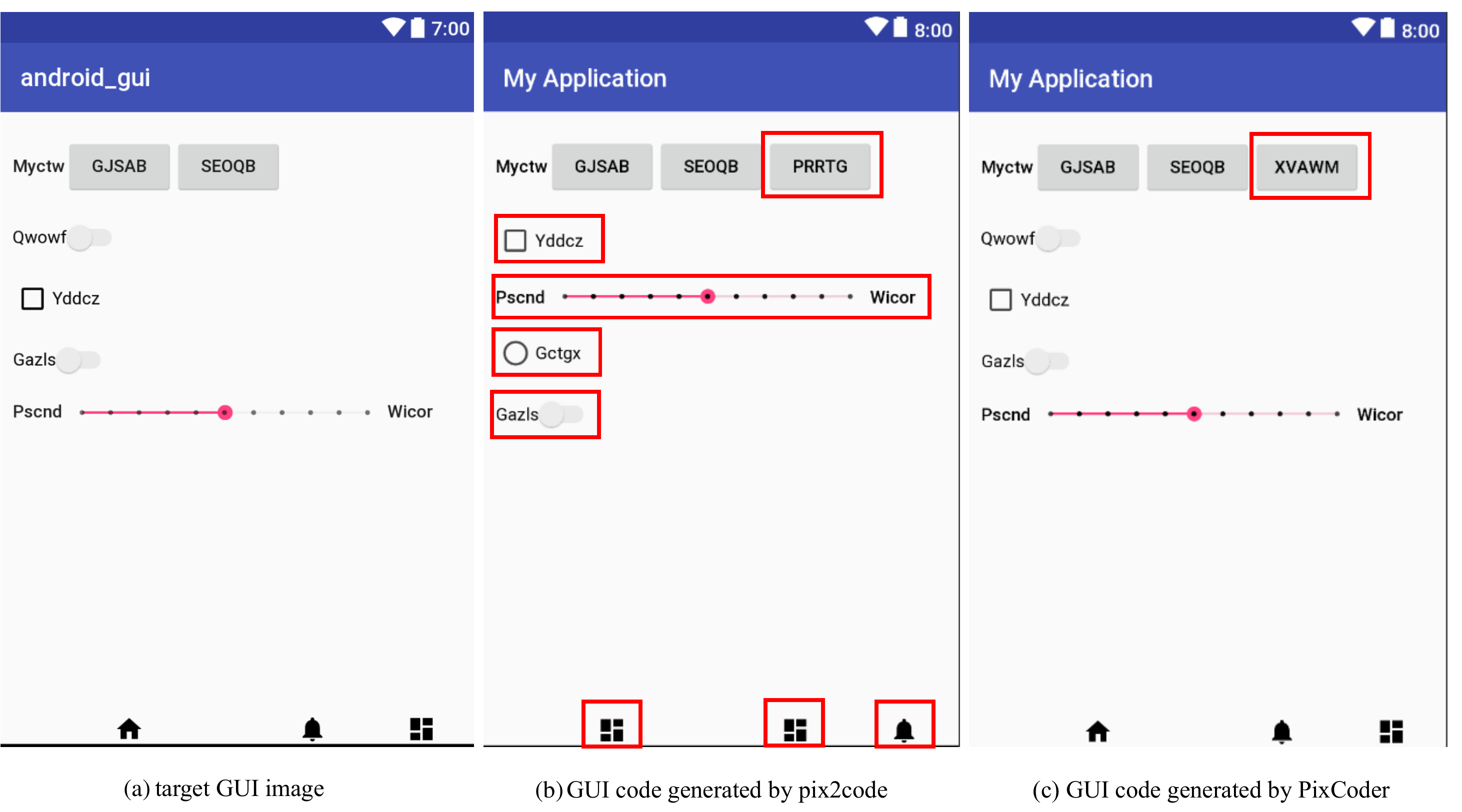}
  \caption{The input Android UI image and the results of different models.}
\mvp\mvp
\label{result}
\end{figure}

\section{Conclusion}
In this paper, we present a method called PixCoding for image recognition and classification. PixCoding decomposes the image classification tasks into sub-classification tasks in different regions of the image. This process is an artificially supervised attention mechanism. We artificially design a vector to encode the image. The vector-image pairs guide CNN model to identify specific regions in the image and classify style sheets in these regions. In the end, following some rules, we can get the final classification results by integrating the results of sub-classification tasks. Our experimental results also show that PixCoding has an excellent effect on image recognition and classification tasks.

We apply the PixCoding to automatic programming field and propose PixCoder. PixCoder is an impressive automatic descriptive programming system, which takes an GUI image as input, and then generates the corresponding front-end code. Compared to other automatic descriptive programming systems, PixCoder produces extremely accurate GUI code that is close to human levels.

However, the GUI image in the dataset we used are simpler than the actual GUI image. Therefore, how to make PixCoder still perform excellent in more complex datasets is our future work. Moreover, we also interested in how to apply PixCoding in other image classification tasks.

\newpage
\bibliographystyle{named}
\bibliography{ijcai18}

\end{document}